# TG-LMM: Enhancing Medical Image Segmentation Accuracy through Text-Guided Large Multi-Modal Model

Yihao Zhao[1], Enhao Zhong[1], Cuiyun Yuan[2], Yang Li[2], Man Zhao[2], Chunxia Li[3], Jun Hu[1], Chenbin Liu[2]

1. School of Electronics and Communication Engineering, Sun Yat-sen University, Shenzhen, Guangdong, China

2. National Cancer Center/National Clinical Research Center for Cancer/Cancer Hospital & Shenzhen Hospital, Chinese Academy of Medical Sciences and Peking Union Medical College, Shenzhen, Guangdong, China

3. Faculty of Health Sciences, University of Macau, Macau, China

Corresponding author: Chenbin Liu, chenbin.liu@gmail.com

**Abstract**

We propose TG-LMM (Text-Guided Large Multi-Modal Model), a novel approach that leverages textual descriptions of organs to enhance segmentation accuracy in medical images. Existing medical image segmentation methods face several challenges: current medical automatic segmentation models do not effectively utilize prior knowledge, such as descriptions of organ locations; previous text-visual models focus on identifying the target rather than improving the segmentation accuracy; prior models attempt to use prior knowledge to enhance accuracy but do not incorporate pre-trained models. To address these issues, TG-LMM integrates prior knowledge, specifically expert descriptions of the spatial locations of organs, into the segmentation process. Our model utilizes pre-trained image and text encoders to reduce the number of training parameters and accelerate the training process. Additionally, we designed a comprehensive image-text information fusion structure to ensure thorough integration of the two modalities of data. We evaluated TG-LMM on three authoritative medical image datasets, encompassing the segmentation of various parts of the human body. Our method demonstrated superior performance compared to existing approaches, such as MedSAM, SAM and nnUnet.

## 1. Introduction

Segmentation is a pivotal task in medical imaging analysis, involving the identification and delineation of regions of interest (ROI) within various medical images, such as organs, lesions, and tissues.(Isensee, Jaeger, Kohl, Petersen, & Maier-Hein, 2021; Ronneberger, Fischer, & Brox, 2015) Accurate segmentation is essential for numerous clinical applications, including disease diagnosis, treatment planning, and monitoring disease progression. Traditionally, manual segmentation has been the gold standard for delineating anatomical structures and pathological regions.(De Fauw et al., 2018; Ouyang et al., 2020) However, this method is time-consuming, labor-intensive, and demands a high level of expertise. Semi-automatic or fully automatic segmentation techniques can significantly reduce the time and effort needed, enhance consistency, and facilitate the analysis of large-scale datasets.

Deep learning-based models have shown considerable promise in medical image segmentation due to their ability to learn intricate image features and deliver precise segmentation results across a wide array of tasks, from segmenting specific anatomical structures to identifying pathological regions. Automatic segmentation methods can be broadly categorized into convolution-based methods and transformer-based methods. Convolution-based methods, such as U-Net(Ronneberger et al., 2015), 3D U-Net (Çiçek, Abdulkadir, Lienkamp, Brox, & Ronneberger, 2016), V-Net(Milletari, Navab, & Ahmadi, 2016), Seg-Net(Badrinarayanan, Kendall, Cipolla, & intelligence, 2017), DeepLab(L.-C. Chen, Zhu, Papandreou, Schroff, & Adam, 2018), VoxResNet(H. Chen, Dou, Yu, Qin, & Heng, 2018) and Mask RCNN(K. He, Gkioxari, Dollár, & Girshick, 2017), have achieved remarkable success in accurately segmenting target regions since their inception in 2016. However, a significant limitation of many current medical image segmentation models is their task-specific design. These models are typically developed and trained for a particular segmentation task, and their performance can deteriorate considerably when applied to new tasks or different types of imaging data.

The introduction of transformers into image analysis in 2020 by Alexey et al.(Dosovitskiy et al., 2020) marked a significant advancement, offering improved generality. Subsequently, Nicolas et al.(Carion et al., 2020) designed a transformer network for object detection, and Alexander et al. (Kirillov et al., 2023) developed the first transformer-based network, SAM, for pixel-level precise segmentation. This network possesses zero-shot learning capabilities and can segment different objects based on various prompts such as points and bounding boxes. Compared to traditional convolutional segmentation networks, it demonstrates greater generalization and transferability. Jun et al.(Ma et al., 2024) adapted the SAM model for the medical imaging field, addressing the unique challenges posed by medical images, such as the complex anatomical descriptions required for organs, tumors, and blood vessels, and the higher segmentation accuracy needed compared to natural images.

In the realm of text-guided image segmentation, numerous approaches have been explored to teach machines to comprehend the visual world through natural language.(Bao et al., 2022; D. Li, Li, & Hoi, 2024; J. Li, Li, Savarese, & Hoi, 2023; J. Li, Li, Xiong, & Hoi, 2022; L. H. Li, Yatskar, Yin, Hsieh, & Chang, 2019; Lin et al., 2021; Radford et al., 2021; W. Wang et al., 2023) Most models are designed for image-text dialogue and image-text matching.(Bao et al., 2022; J. Li et al., 2023; J. Li et al., 2022; L. H. Li et al., 2019; Radford et al., 2021) With the rise of generative artificial intelligence, some models have also been used for text-guided image generation.(D. Li et al., 2024; Ramesh et al., 2021; W. Wang et al., 2023) The SAM model was the first to introduce textual information into segmentation, employing a two-stage approach that generates multiple possible independent object masks and then uses an image classification model to identify the object that best matches the given text.(H. Wang et al., 2024) However, this approach does not improve accuracy because textual information is not involved in the mask generation process. Similarly, Tomar et al. (Tomar, Jha, Bagci, & Ali, 2022) attempted to use a series of cross-attention mechanisms for text-guided medical image segmentation, focusing was on segmenting different types of medical images rather than improving segmentation accuracy. Li et al. (Z. Li et al., 2023) used text to enhance segmentation accuracy, achieving good results with a model consisting of a cross-structure of CNN and transformer. With the rise of pre-trained multifunctional large models in recent years(Kojima, Gu, Reid, Matsuo, & Iwasawa, 2022), retraining a network is not the best approach.

In summary, previous methods for text-guided delineation have the following issues:

**(1)Current medical automatic segmentation models do not effectively utilize prior knowledge, such as descriptions of organ locations.**
**(2)Most text-visual models aim to identify the target while segmenting, rather than improving accuracy.**
**(3)Some models attempt to use prior knowledge to enhance accuracy but do not incorporate pre-trained models.**

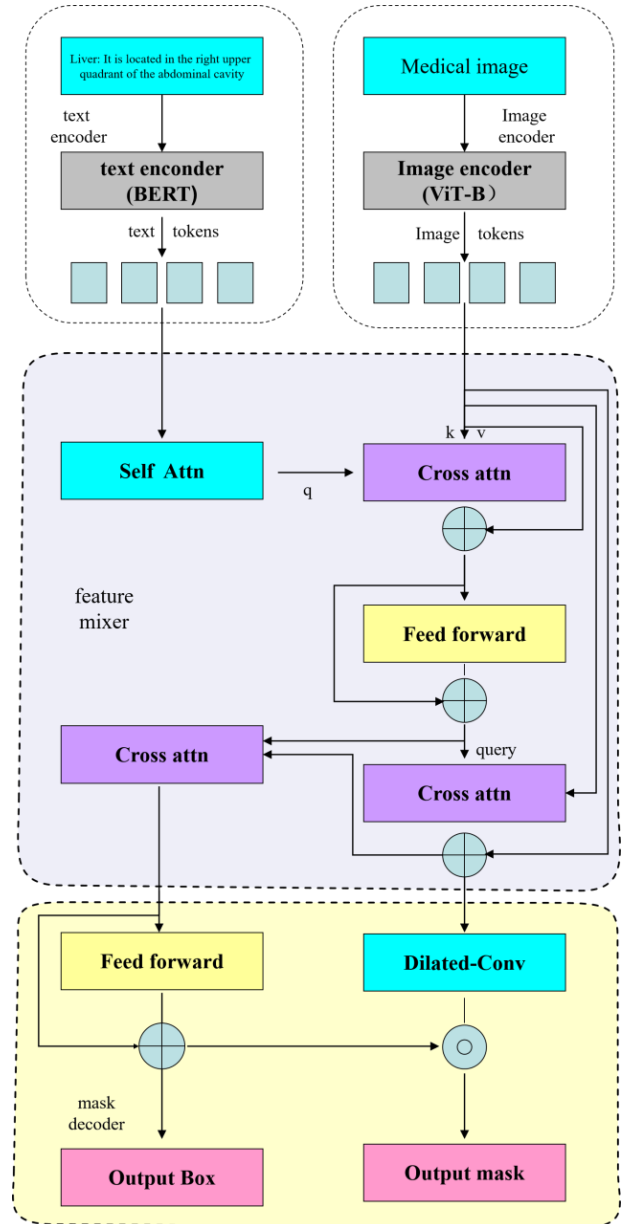

Figure 1. The structure of our model includes a transformer-based (Dosovitskiy et al., 2020) image encoder, a BERT-based text encoder (Radford et al., 2021), a query-based feature mixer (Carion et al., 2020), and a mask decoder.

By addressing these issues, our research focus on developing more robust and generalizable segmentation models that leverage prior knowledge and pre-trained models to improve accuracy and efficiency in medical image segmentation.

## 2. Related works

**Object detection & segmentation.** In the past, researchers have proposed a series of convolution-based methods to achieve object detection and segmentation.(Duan et al., 2019; Girshick, Donahue, Darrell, & Malik, 2014; K. He et al., 2017; Liu et al., 2016; Radford et al., 2021; Radford et al., 2019; Redmon, Divvala, Girshick, & Farhadi, 2016; Ren, He, Girshick, & Sun, 2015) With the advancement of transformers, researchers have proposed a series of transformer-based methods, which are capable of achieving interactive segmentation.(Carion et al., 2020; Kirillov et al., 2023; Ma et al., 2024; H. Wang et al., 2024; W. Wang et al., 2023) Unlike previous work, our goal is not to guide segmentation through text or perform image-text matching after segmentation. Instead, we aim to use text to enhance the accuracy of segmentation.

**Multimodal learning,** Multimodal learning has been applied to image-text matching problems. Unlike traditional convolutional models that rely on classification methods, multimodal learning offers significantly stronger generalization capabilities. (Bao et al., 2022; D. Li et al., 2024; J. Li et al., 2023; J. Li et al., 2022; Radford et al., 2021) It have also been utilized for video and image generation. (D. Li et al., 2024; Ramesh et al., 2021; W. Wang et al., 2023) In addition, it has also been applied to both medical and natural image segmentation(Kirillov et al., 2023; Ma et al., 2024; H. Wang et al., 2024; W. Wang et al., 2023), where it accepts points and bounding boxes as prompt inputs, demonstrating strong cross-modal capabilities. This allows the models to effectively integrate different types of input data, leading to more precise and flexible segmentation results.

## 3. Method

### 3.1 The whole workflow of our model

The structure of our model is shown in figure 1. Our model comprises four main components: a transformer-based (Dosovitskiy et al., 2020) image encoder, a BERT-based text encoder (Radford et al., 2021), a query-based feature mixer (Carion et al., 2020), and a mask decoder. During training, the parameters of the image encoder and text encoder were frozen, while the feature mixer and mask decoder were trained. The total number of parameters is 218 million, with 5.6 million trainable parameters. Due to the limited availability of medical image datasets, especially the textual description, training the entire model would lead to severe overfitting. Therefore, we initialized the image encoder and text encoder with weights from pre-trained models on natural images. (Radford et al., 2021) The model's input is a slice of CT image, along with a description provided by a clinical expert. This description includes the anatomical position of the organ and its spatial relationship with other organs (supplemental material). The output is a precise segmentation mask.

### 3.2 Image encoder

Considering computational complexity, we used the ViT base model as the image encoder. It consists of 12 transformer layers, each containing a 12-head multi-head attention block with 768 hidden dimensions and an MLP with 3072 hidden dimensions. The input CT image dimensions were resampled to 1024x1024x3 and then downsampled by a factor of 16, resulting in a 16x16x3 sequence. The image encoder extracted features from the image, converting it into a 768-dimensional vector. The formalized computation process is as follows:

$$F_{im} = E_{image}(I)$$

where I is the input image, $E_{image}()$ is the image encoder, and $F_{im}$ is the image feature vector

### 3.3 Text encoder

We chose a transformer-based encoder as the text encoder.(Radford et al., 2021; Vaswani et al., 2017) It consists of 12 layers, each with 512 hidden variables and 8 attention heads. The text sequence is enclosed with [SOS] and [EOS] tokens, and the activations of the highest layer of the transformer at the [EOS] token were used as the feature representation of the text. This representation is layer-normalized and then linearly projected into the multi-modal embedding space. Masked self-attention is employed in the text encoder to maintain the capability of adding language modeling as an auxiliary objective, though this aspect is reserved for future exploration.

$$F_{text} = E_{text}(T)$$
$$T = E_{text}(T)$$

Where $T$ is the input text, $E_{text}()$ is the text encoder, and $F_{text}$ is the text feature vector

### 3.4 Feature mixer

Inspired by SAM(Kirillov et al., 2023), we designed a query-based image-text feature fusion module. It includes a self-attention module(Vaswani et al., 2017), two cross-attention modules, and a feed-forward neural network(Pinkus, 1999). The dimension of the attention layer is set to 256, and the hidden dimension of the MLP is set to 2048, respectively. After extracting the image and text information $F_{im}$ and $F_{text}$ by the Image encoder and Text encoder, the text vector first passes through a self-attention module. Then, it acts as a query in a cross-attention module with the image vector, followed by an MLP, and is output as the fused text vector $F_{fused\ text}$. This fused text vector then acts as a query in a cross-attention module with $F_{im}$, resulting in the fused image vector output $F_{fused\ im}$. Each computation in the above process is accompanied by skip connections to avoid the gradient vanishing problem.(K. He, Zhang, Ren, & Sun, 2016) The formalized computation process is as follows:

$$F_{text,1} = F_{text} + attn(F_{text}, F_{text})$$
$$F_{text,2} = F_{im} + cross\_attn(F_{text,1}, F_{im})$$
$$F_{fused\ text} = F_{text,2} + Feed\_Forward(F_{text,2})$$
$$F_{fused\ im} = F_{im} + cross\_attn(F_{fused\ text}, F_{im})$$

### 3.5 Mask decoder

The mask decoder's function is to decode pixel-level segmentation masks from the fused image and text vectors. It consists of two consecutive dilated convolutions, a bi-directional transformer, and two MLPs. Each dilated convolution reduces the number of channels by half and double the size of the feature map. The previously output text feature vector and the output feature map go through a bi-directional attention structure for further fusion, resulting in the output feature map. Subsequently, two MLPs decode this map into a pixel-level precise segmentation mask and a bounding box to accelerate convergence. The formalized computation process is as follows:

$$F_1 = \text{Dilated\_Conv}(F_{fused\ im})$$
$$F_2 = \text{Dilated\_Conv}(F_1)$$
$$F_3 = \text{cross\_atten}(F_{fused\ im}, F_{fused\ text})$$
$$\text{Mask} = \text{MLP}_{\text{mask}}(F_3) * F_2$$
$$\text{Bounding\_Box} = \text{MLP}_{\text{bbox}}(F_3)$$

Here, $F_{fused\ im}$ is the fused image vector, and $F_{fused\ text}$ is the fused text vector. The outputs are the segmentation mask and the bounding box. The feature mixer are stacked four times to fully blend the image and text features. During the training process, the feature mixer and the mask decoder were optimized.

### 3.6 Loss function

We utilized the unweighted sum of cross-entropy loss and dice loss as the final loss function, as it has proven to be robust across various medical image segmentation tasks. The binary cross-entropy loss is defined by the following formula:

$$L_{\text{BCE}}(AGC, GT) = -\frac{1}{N}\sum_{i=1}^{N}[g_i \log a_i + (1-g_i)\log(1-a_i)]$$

The dice loss is defined by:

$$L_{\text{dice}}(AGC, GT) = 1 - \frac{2\sum_{i=1}^{N} g_i a_i}{\sum_{i=1}^{N}(g_i)^2 + \sum_{i=1}^{N}(a_i)^2}$$

Where $AGC$ and $GT$ denote the segmentation result and ground truth, respectively. Let $a_i$ and $g_i$ present the predicted segmentation and ground truth of voxel $i$, respectively, and $N$ be the number of voxels in the image $I$. The final loss L is defined by:

$$L(AGC, GT) = L_{\text{BCE}}(AGC, GT) + L_{\text{dice}}(AGC, GT)$$

## 4. Experiment

### 4.1 Implementation Details

Our model consists of four key components: an image encoder, a text encoder, a text-image feature mixer, and a mask decoder. The image encoder utilizes a pretrained ViT-base, (Dosovitskiy et al., 2020) while the text encoder employs a pretrained BERT-based model. (Devlin, Chang, Lee, & Toutanova, 2018) The text query tokens (T) have a length of 8, and the image tokens (I) have a length of 6. These modules were optimized for generating region captions. The model has a total number of 218 million parameters, with 5.6 million trainable parameters. Training was conducted on a server equipped with eight A100 40GB GPUs, using approximately 10,000 images and 100,000 high-quality masks. The batch size was set to 8, and the model was trained over 150 epochs, taking about 20 hours to complete. Further details can be found in the supplementary materials table A

### 4.2 Dataset

We used data from the FLARE(Ma et al., 2023), SegTHOR(T. He, Hu, Song, Guo, & Yi, 2020), and MSD(Antonelli et al., 2022) datasets for training and testing. These datasets include CT images and high-quality masks annotated by senior physicians, identifying organs at risk such as the liver, heart, brain tumor, hippocampus, prostate, lung, pancreas, hepatic vessel, spleen, colon, right kidney, spleen, aorta, inferior vena cava, adrenal gland, gallbladder, esophagus, stomach, duodenum, left kidney, and trachea. The diverse distribution of these organs across various parts of the human body ensures the generalizability of our model. To enhance the segmentation process, we used descriptive language from authoritative medical texts regarding the locations of these organs as text input.(Alessandrino et al., 2019; Biga et al., 2020; Collins, Munoz, Patel, Loukas, & Tubbs, 2014; Demetriades, Inaba, & Velmahos, 2020; Drenckhahn & Waschke, 2020; Moore, Dalley, & Agur, 2014; Neil Granger, Holm, & Kvietys, 2011; Rizvi, Wehrle, & Law, 2023; Singh & Bolla, 2019; Standring et al., 2005) The complete descriptions of the organs can be found in the supplementary materials table B.

### 4.3 Evaluation metrics

We adhered to the guidelines in Metrics Reloaded(Maier-Hein et al., 2024), employing the Dice Similarity Coefficient (*DSC*) (Sorensen, 1948) Hausdorff distance (*HD*)(Blumberg, 1920), and average surface distance (*ASD*)(Heimann et al., 2009) to quantitatively assess the segmentation outcomes. The DSC is a region-based segmentation metric designed to evaluate the overlap between expert annotation masks and segmentation results. It is defined by the following formula:

$$\text{DSC}(GT, AGC) = \frac{2|GT \cap AGC|}{|GT| + |AGC|}$$

where GT is the ground truth and AGC is the automatically generated contours.

$HD_{95}$ and $ASD$ are boundary-based metrics to evaluate the boundary consensus between expert annotation masks and segmentation results at a given tolerance, which is defined by:

$$d(X \to Y) = \max_{x \in X} \min_{y \in Y}(d_{\square}^{X \to Y})$$

$HD_{95}(GT, AGC) = max_{95\%}(d(GT \to AGC), d(AGC \to GT))$

where d is the one-sided Euclidean distance from point set X to point set Y. $HD_{95}$ is the longest bidirectional distance between the ground truth and automatically generated contours at the 95$^{th}$ percentile.

$$ASD(GT, AGC) = \frac{1}{N_{GT} + N_{AGC}}$$

$$\left( \sum_{x \in GT} \min_{y \in AGC} \|x - S(AGC)\| + \sum_{y \in AGC} \min_{x \in GT} \|y - S(GT)\| \right)$$

where $N_{GT}$ and $N_{AGC}$ are the number of the pixels in the contour of ground truth and automatically generated contours respectively. $\|\cdot\|$ denotes the Euclidean distance. $S(\cdot)$ denote the point set of surface voxels.

### 4.4 Comparison with other methods

To statistically analyze and compare the performance of the four methods mentioned (MedSAM(Ma et al., 2024), SAM(Kirillov et al., 2023), and U-Net(Isensee et al., 2021) specialist models), we used the Wilcoxon signed-rank test. This non-parametric test is ideal for comparing paired samples, especially when the data does not satisfy the assumptions of a normal distribution. Through this analysis, we determined whether any of the methods demonstrated statistically superior segmentation performance compared to the others, offering valuable insights into the comparative effectiveness of the evaluated methods.

## 5. Result

### 5.1 Statistical results

Table 1 demonstrates the superiority of our method compared to other approaches. It provides a comparative analysis of performance metrics across different datasets, highlighting the improved accuracy and robustness of our proposed model. Compared to the previous state-of-the-art method, MedSAM, our approach shows only a slight improvement in the region-based segmentation metric, with

Table 1. The comparison results of four different methods, including our approach, across all datasets used in this experiment, as well as the individual results for the FLARE, MSD, and SegTHOR datasets, are summarized below. The data format is presented as mean ± standard deviation. The best result in each metric is bolded.

|  |  | ALL | FLARE | SegTHOR | MSD |
|---|---|---|---|---|---|
| nnUnet | DSC | 0.946±0.033 | 0.975±0.020 | 0.953±0.021 | 0.900±0.038 |
|  | $HD_{95}$ | 90.25±125.22 | 72.58±131.88 | 19.96±31.45 | 145.11±116.65 |
|  | ASD | 26.21±27.88 | 26.22±41.55 | 6.11±7.88 | 41.33±35.18 |
| SAM | DSC | 0.568±0.162 | 0.410±0.16 | 0.486±0.151 | 0.860±0.079 |
|  | $HD_{95}$ | 196.25±202.66 | 136.89±198.53 | 37.19±52.89 | 170.12±112.43 |
|  | ASD | 48.55±54.11 | 34.21±76.16 | 9.78±12.87 | 42.67±45.22 |
| MedSAM | DSC | 0.953±0.029 | 0.980±0.015 | 0.956±0.023 | 0.908±0.04 |
|  | $HD_{95}$ | 98.45±120.74 | 70.45±120.74 | 18.17±33.85 | 131.22±126.39 |
|  | ASD | 24.41±33.35 | 20.51±40.21 | 4.31±9.72 | 34.87±38.36 |
| TG-LMM (ours) | DSC | 0.953±0.029 | 0.98±0.015 | **0.970±0.014** | **0.910±0.04** |
|  | $HD_{95}$ | **69.22±110.70** | **44.05±82.47** | **6.205±23.53** | **101.79±129.93** |
|  | ASD | **15.21±28.92** | **9.62±20.11** | **1.33±5.93** | **22.53±35.52** |

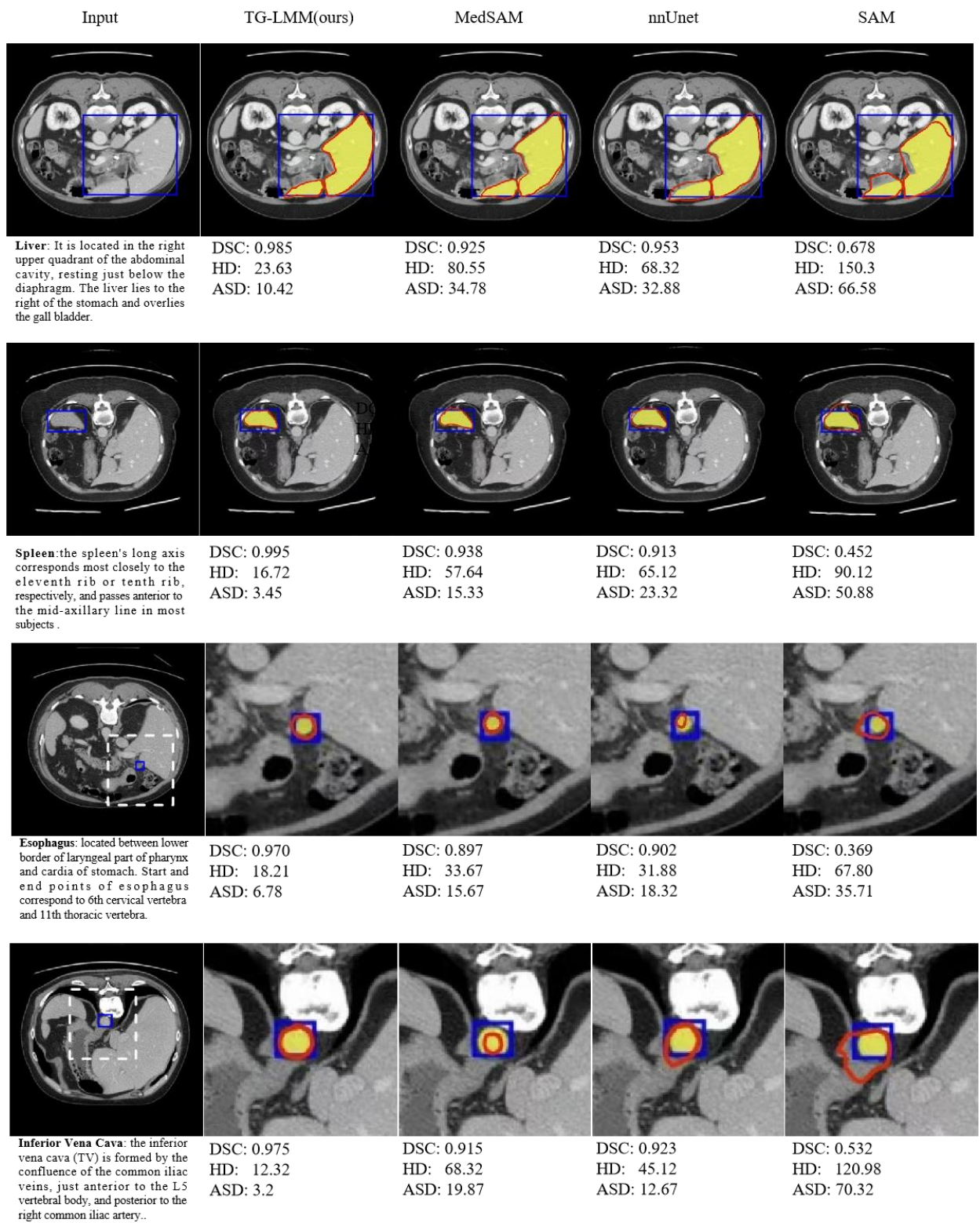

Figure 2. The segmentation examples generated by our method, alongside comparisons with other methods, are presented. The blue bounding box denotes the input prompt, the yellow filled area represents the ground truth, and the red contour indicates the automatically generated segmentation.

However, for boundary-based metrics, our method demonstrates significant enhancements: the mean $HD_{95}$ decreased by 29.22, with a variance reduction of 10.04; the mean ASD decreased by 9.2, with a variance reduction of 4.43. These results indicate that our approach is highly effective in improving the precision of boundary delineation. The statistical results for each organ individually can be found in the supplementary materials table C. Some representative contouring examples are displayed in Figure 2.

**5.2 Ablation experiment**
We designed an ablation experiment to evaluate the effectiveness of inputting authoritative text descriptions. To test the impact of inputting authoritative descriptions on segmentation results, we retrained and tested the model using three different types of input: complex descriptions (our method), simple descriptions (only the organ names), and no descriptions. This allowed us to assess how varying levels of descriptive detail affect the model's segmentation performance. The results of this experiment are presented in Table 2.

Table 2 The ablation of complexity of descriptions

| descriptions | DSC | HD | ASD |
|---|---|---|---|
| no | 0.953 | 98.45 | 24.41 |
| simple | 0.953 | 96.56 | 28.81 |
| complex(ours) | 0.953 | 69.22 | 15.21 |

The results show that simply adding text to label the organ does not significantly improve the contour delineation. However, using authoritative textual descriptions significantly enhances the accuracy of the automated contouring, demonstrating the value of detailed expert input in refining segmentation precision. This suggests that in the future, we could potentially develop a set of authoritative guidelines that are better understood by models, specifically designed to enhance automatic contouring. These guidelines would leverage expert knowledge to improve the precision and reliability of automated segmentation processes.

# 6. Conclusion and discussion

**6.1 Conclusion.** We developed an automatic segmentation model for medical images that utilizes authoritative descriptions of organs as prompts to achieve high-precision segmentation. A key component of our model is a feature mixer that fully integrates text input with visual input, thereby enhancing segmentation performance. Unlike previous models(Carion et al., 2020; Kirillov et al., 2023), our objective is not to locate the segmentation target based on text but rather to improve segmentation accuracy through detailed textual descriptions. Experimental results indicate that using authoritative descriptions significantly improve organ contour segmentation, leading to substantial improvements in boundary-based metrics across several authoritative datasets.(Antonelli et al., 2022; T. He et al., 2020; Ma et al., 2023) Additionally, we compared our method with approaches that use simple text input and no text input, further validating the effectiveness of our approach.

**6.2 Discussion**
**Effectiveness on sequential organs.** In 2D image segmentation, sequential organs, such as the esophagus, present a significant challenge due to their small size and in images. Our method demonstrates substantial improvements over other approaches for segmenting this type of organs (Figure 2). Specifically, for the esophagus, the DSC improved by an average of 0.42, the HD decreased by 4.24, and ASD decreased by 2.23. Similarly, for the inferior vena cava, the DSC improved by 0.04, the HD decreased by 32.17, and the ASD decreased by 6.11, compared to MedSAM.(Ma et al., 2024) There results not only further validates the effectiveness of our method but also encourage further exploration of more effective and computationally efficient problem descriptions to enhance segmentation accuracy for the sequential organs.

**Improvement in Robustness** The quality of automated contouring has long been a significant challenge for researchers.(Zhao et al., 2024) Although existing models can achieve contouring accuracy comparable to that of expert physicians in many scenarios, they often struggle with poor segmentation accuracy for certain organs. Our approach demonstrates smaller variance across various segmentation metrics compared to traditional methods, indicating enhanced robustness. While the average improvements in metrics such as DSC may not be substantial, the consistent performance of our method highlights its significant potential for industrial applications. This consistency ensures reliable outcomes across diverse cases, making it a valuable tool for practical use.

**Limitations** Due to the relatively small size of the dataset in this experiment, training too many parameters could lead to overfitting. Consequently, we did not train the text encoder and image encoder. Additionally, the application of large text models in the medical field is still in its early stages, which may result in suboptimal efficiency when using a pretrained text encoder to extract descriptive information. Furthermore, the limited availability of medical image datasets, which often contain only specific organs while leaving other parts unsegmented, renders our model incomplete. We hope that as medical imaging technology advances, more comprehensive datasets will become available, allowing us to train a model capable of understanding and segmenting all objects within the human body. In the field of medical imaging, there are numerous atlas guides that are recognized by authoritative experts as highly representative segmented images, often accompanied by detailed instructions. Our method has only considered using textual information as prompts, without exploring the potential of incorporating additional modalities as prompts.

**Future work** In the future, we aim to integrate various authoritative guidelines and leverage knowledge graph technology to achieve higher precision and broader domain medical image segmentation. Our goal is to develop a system capable of accepting new guidelines and tasks without additional training, delivering high-quality segmentation results similar to the operation of current large language models.(Kojima et al., 2022) We also encourage collaboration among researchers in the field to provide richer datasets for multimodal medical tasks, with the aim of developing more intelligent medical models. By fostering a collaborative environment and utilizing advanced technologies, we hope to push the boundaries of medical image segmentation and improve the overall quality of automated medical diagnostics.

# Supplemental material
Yihao Zhao[1], Enhao Zhong[1], Cuiyun Yuan[2], Yang Li[2], Man Zhao[2], Chunxia Li[3], Jun Hu[1], Chenbin Liu[2]

1. School of Electronics and Communication Engineering, Sun Yat-sen University, Shenzhen, Guangdong, China

2. National Cancer Center/National Clinical Research Center for Cancer/Cancer Hospital & Shenzhen Hospital, Chinese Academy of Medical Sciences and Peking Union Medical College, Shenzhen, Guangdong, China

3. Faculty of Health Sciences, University of Macau, Macau, China

Corresponding author: Chenbin Liu, chenbin.liu@gmail.com

### A. Implementation Details

Tab. A presents full implementation details of our method.

**Table A** Implementation Details

| | |
|---|---|
| **Optimization** | |
| Optimizer | AdamW (0.9, 0.999) |
| Learning Rate | 0.0001 |
| LR Decay Ratio | 0 |
| LR Decay | cosine |
| Weight Decay | 0.0001 |
| Warmup ratio | 0.3333 |
| Warmup steps | 1 |
| Gradient Clipping | 1.0 |
| **Data Epoch** | |
| Batch Size | 16 |
| Steps | 150 |
| Img | 47,301 |
| mask | 108,036 |
| GPU type | A100-40G |
| GPU num | 8 |
| **Model Details** | |
| Input | a) 1024x1024 Long side: 1024 Short side: padding b) Large Scale Jitter c) Horizontal Flip |
| Loss | a) Cross Entropy Loss b) dice Loss |
| Text encoder | BERT |
| Parameters | 216993328 |
| Trainable parameters | 4255204 |

### B. Complete descriptions

Tab. B presents a complete description of different organs. These descriptions are all derived from authoritative textbooks or guidelines.(Alessandrino et al., 2019; Biga et al., 2020; Collins, Munoz, Patel, Loukas, & Tubbs, 2014; Demetriades, Inaba, & Velmahos, 2020; Drenckhahn & Waschke, 2020; Moore, Dalley, & Agur, 2014; Neil Granger, Holm, & Kvietys, 2011; Rizvi, Wehrle, & Law, 2023; Singh & Bolla, 2019; Standring et al., 2005)

**Table B**

| Organs | Descriptions |
|---|---|
| liver | The liver is located inferior to the diaphragm, primarily occupying the right upper quadrant of the abdomen, from the fifth intercostal space to the right costal margin. It is mostly intraperitoneal, with a bare area on its superior posterior aspect adjacent to the diaphragm and inferior vena cava. The liver consists of four lobes—right, left, caudate, and quadrate—and is closely associated with the gallbladder and right kidney, divided by the falciform ligament(Alessandrino et al., 2019). |
| left kidney | The vertebral limits of the left kidney are T12–L3 or L4, while those for the right kidney are L1–L4 (overall range upper T11 to lower L5). The upper poles of both kidneys lie anterior to rib 12, and they lie anterior to the rib 11 in 30% (left) and 10% (right) of subjects(Standring et al., 2005). The left kidney typically lies between the T12 and L3 or L4 vertebrae, with its upper pole positioned anterior to rib 12. In some cases, it may be lower than the right kidney, which usually sits 2 cm higher. The renal hilum is generally located at L1/2 or L2. The left kidney measures about 11.5 cm in length and is slightly longer than the right, with a width of 5–6 cm(Standring et al., 2005). |
| spleen | The spleen's long axis corresponds most closely to the eleventh rib or tenth rib, respectively, and passes anterior to the mid-axillary line in most subjects. The spleen extends from a point about 5 cm to the left of the posterior midline at the level of the eleventh thoracic spine and extends about 3 cm anterior to the mid-axillary line. Its normal size approximates roughly to that of the individual's clenched fist(Standring et al., 2005). |
| pancreas | The surface projection of the head of the pancreas lies within the duodenal curve on the right side of the second lumbar vertebra; the neck lies in the transpyloric plane, level with the L1/2 intervertebral disc; and the body passes obliquely up and to the left towards the spleen, lying slightly above the transpyloric plane, near the tail(Standring et al., 2005). |
| aorta | The aorta has many sections, including the aortic root, a section that attaches to the heart. This is the widest part of the aorta; ascending aorta: Upward curve that occurs shortly after the aorta leaves the heart. aortic arch: Curved segment that gives the aorta its cane-like shape. It bridges the ascending and descending aorta. descending aorta: Long, straight segment that runs from your chest (thoracic aorta) to your abdominal area (abdominal aorta)(Collins et al., 2014). |
| inferior vena cava | The inferior vena cava (IVC) forms from the common iliac veins at L5, posterior to the right common iliac artery. It ascends to the right of the vertebral bodies, entering the thorax at T8, where the right crus of the diaphragm separates it from the aorta. Typically, a 1 cm suprahepatic IVC segment between the liver and diaphragm allows cross-clamping for surgical access(Demetriades et al., 2020). |
| right adrenal glands | The right adrenal glands are glandular and neuroendocrine tissue adhering to the top of the right kidneys by a fibrous capsule. Both adrenal glands sit atop the kidneys and are composed of an outer cortex and an inner medulla, all surrounded by a connective tissue capsule(Biga et al., 2020). |
| left adrenal glands | The left adrenal glands are glandular and neuroendocrine tissue adhering to the top of the left kidneys by a fibrous capsule. Both adrenal glands sit atop the kidneys and are composed of an outer cortex and an inner medulla, all surrounded by a connective tissue capsule(Biga et al., 2020). |
| gallbladder | The fundus of the gallbladder is commonly identified with the tip of the ninth costal cartilage (in the transpyloric plane), near the junction of the linea semilunaris with the costal margin. Recent data have shown that the fundus lies in the transpyloric plane |
| | in approximately one-third of supine individuals and below the plane in most others(Standring et al., 2005). |
| esophagus | The esophagus is located between lower border of laryngeal part of pharynx and cardia of stomach. The start and end points of esophagus correspond to the 6th cervical vertebra and the 11th thoracic vertebra topographically, and the gastroesophageal junction corresponds to the xiphoid process of the sternum(Rizvi et al., 2023). |
| Stomach | The gastro-oesophageal junction lies to the left of the midline, posterior to the left seventh costal cartilage, at the level of the eleventh thoracic vertebra (range upper T10 to L1/2), with the level being lower in females and higher in the obese. The stomach lies in a curve within the left hypochondrium and epigastrium, although, when distended, it may lie as far down as the umbilical or suprapubic regions(Standring et al., 2005). |
| duodenum | The first part of the duodenum sometimes ascends above the transpyloric plane; the second part usually lies just to the right of the midline, alongside the second and third lumbar vertebrae; the third part usually crosses the midline at the level of the third lumbar vertebra; and the fourth part ascends to the left of the second lumbar vertebra, reaching the transpyloric plane in the region of the lower border of the first lumbar vertebra(Standring et al., 2005). |
| spinal cord | The spinal cord occupies the superior two-thirds of the vertebral canal. It is continuous cranially with the medulla oblongata, just below the level of the foramen magnum. The spinal cord terminates on average at the level of the middle third of the body of the first lumbar vertebra. The spinal cord varies in transverse width, gradually tapering craniocaudally, except at the levels of enlargements(Standring et al., 2005). |
| heart | The upper border slopes from the second left costal cartilage to the third right costal cartilage. The right border is a curved line, convex to the right, running from the third to the sixth right costal cartilage, usually 1-2 cm lateral to the sternal edge. The inferior or acute border runs leftwards from the sixth right costal cartilage to the cardiac apex, located approximately 9 cm lateral to the midline, in the left fifth intercostal space(Standring et al., 2005). |
| Left lung | The left subclavian artery and left brachiocephalic vein are adjacent to the medial surface of the apex of the left lung. The basal surface is semilunar and concave, moulded on the superior surface of the diaphragm, which separates each lung from the corresponding lobe of the liver, and the left lung from the gastric fundus and spleen. The mediastinal area is deeply concave, which is much larger and deeper on the left lung where the heart projects more to the left of the median plane(Standring et al., 2005). |
| Right lung | Scalenus medius is lateral and the brachiocephalic trunk, right brachiocephalic vein, and trachea are adjacent to the medial surface of the apex of the right lung. The basal surface is semilunar and concave, moulded on the superior surface of the diaphragm, which separates each lung from the corresponding lobe of the liver, and the left lung from the gastric fundus and spleen(Standring et al., 2005). |
| colon | The ascending colon extends from the cecum to the hepatic flexure on the right, lying behind the peritoneum. The transverse colon runs horizontally from the hepatic to splenic flexure, covered by peritoneum and attached by the transverse mesocolon. The descending colon extends from the splenic flexure to the sigmoid colon on the left, also retroperitoneal. The sigmoid colon connects to the rectum, attached by the sigmoid mesocolon, transitioning at the third sacral vertebra(Drenckhahn & Waschke, 2020). |
| hepatic vessel | Hepatic Portal Vein is Located in the upper right quadrant of the abdomen, it originates behind the neck of the pancreas and enters the liver through the porta hepatis. Hepatic Veins: These veins (right, middle, and left) are located at the top of the liver, connecting to the inferior vena cava(Neil Granger et al., 2011). |
| prostate | The prostate lies directly inferior to the bladder and wraps around the proximal urethra in the lesser pelvis. Anteriorly, it is posterior to the pubic symphysis, separated by a pad of fat and venous plexus. Posteriorly, it is close to the rectum and is separated by |
| | fascia of Denonvilliers.The external urethral sphincter muscle is beneath the prostate. Laterally, the gland is related to the levator ani muscle of the pelvic floor covered by endopelvic fascia(Singh & Bolla, 2019). |
| trachea | The trachea extends from the lower edge of the larynx (cricoid cartilage) at the level of the 6th cervical vertebra (C6) to the carina at the 4th-5th thoracic vertebra (T4-T5). It's located in the anterior neck and upper chest, anterior to the esophagus. The trachea begins in a midline position but inclines slightly to the right as it descends. It has a cervical portion in the neck and a thoracic portion in the superior mediastinum(Moore et al., 2014). |

**C. Complete result**

Table C.A presents the Dice Similarity Coefficient (DSC) of various organs in this experiment, along with comparison results with other methods. (Isensee, Jaeger, Kohl, Petersen, & Maier-Hein, 2021; Kirillov et al., 2023; Ma et al., 2024)Table C.B shows the Hausdorff Distance at $95^{th}$ percentile ($HD_{95}$) of the same organs, including comparative results with other methods. Table C.C displays the Average Surface Distance (ASD) of these organs, also including comparisons with alternative methods. The organs evaluated in this study include the liver, right kidney, spleen, pancreas, aorta, inferior vena cava, right adrenal gland, left adrenal gland, gallbladder, esophagus, heart, left kidney, trachea, hippocampus, stomach, hepatic vessel, and colon.

Table C.A The complete comparison of DSC for four different methods

|  | TG-LMM (ours) | MedSAM | SAM | nnUnet |
|---|---|---|---|---|
| ALL | **0.953±0.029** | 0.953±0.029 | 0.568±0.162 | 0.946±0.033 |
| FLARE | **0.980±0.015** | 0.980±0.015 | 0.410±0.16 | 0.975±0.020 |
| segThor | **0.970±0.014** | 0.956±0.023 | 0.486±0.151 | 0.953±0.021 |
| MSD | **0.910±0.04** | 0.908±0.04 | 0.860±0.079 | 0.900±0.038 |
| liver | **0.974±0.012** | 0.964±0.009 | 0.901± 0.07 | 0.961±0.01 |
| Right kidney | **0.982±0.010** | 0.948±0.03 | 0.490±0.017 | 0.933±0.027 |
| spleen | **0.987±0.007** | 0.986±0.0009 | 0.511±0.018 | 0.973±0.01 |
| Pancreas | **0.885±0.046** | 0.885±0.044 | 0.343±0.16 | 0.863±0.017 |
| Aorta | **0975±0.012** | 0.974±0.012 | 0.807±0.010 | 0.932±0.02 |
| Inferior Vena Cava | **0964±0.0004** | 0.960±0.0004 | 0862±0.11 | 0.928±0.01 |
| Right Adrenal Gland | **0987±0.015** | 0.986±0.017 | 0.100±0.09 | 0.921±0.023 |
| Left Adrenal Gland | **0984±0.014** | 0.983±0.016 | 0.122±0.09 | 0.951±0.021 |
| Gallbladder | **0.987±0.017** | 0.985±0.016 | 0.191±0.15 | 0.961±0.012 |
| Esophagus | **0.942±0.028** | 0.900±0.003 | 0.452±0.157 | 0.881±0.019 |
| heart | **0.981±0.001** | 0.980±0.001 | 0.431±0.16 | 0.941±0.008 |
| Left kidney | 0.988±0.007 | 0.986±0.009 | 0.481±0.16 | **0.989±0.0003** |
| trachea | **0.967±0.019** | 0.969±0.002 | 0.271±0.12 | 0.962±0.005 |
| hippocampus | 0.889±0.02 | **0.900±0.025** | 0.403±0.02 | 0.890±0.03 |
| Stomach | **0.975±0.02** | 0.975±0.02 | 0.423±0.176 | 0.961±0.04 |
| hepatic vessel | **0.912±0.06** | 0.880±0.07 | 0.438±0.18 | 0.892±0.03 |
| colon | **0.815±0.05** | 0.800±0.050 | 0.551±0.190 | 0.810±0.06 |

Table C.B The complete comparison of $HD_{95}$ for four different methods

|  | TG-LMM (ours) | MedSAM | SAM | nnUnet |
|---|---|---|---|---|
| ALL | **69.22±110.70** | 98.45±120.74 | 196.25±202.66 | 90.25±125.22 |
| FLARE | **44.05±82.47** | 70.45±120.74 | 136.89±198.53 | 72.58±131.88 |
| segThor | **6.205±23.53** | 18.17±33.85 | 37.19±52.89 | 19.96±31.45 |
| MSD | **101.79±129.93** | 131.22±126.39 | 170.12±112.43 | 145.11±116.65 |
| liver | **100.13±123.20** | 172.17±123.39 | 177.25±111.12 | 168.33±101.89 |
| Right kidney | **5.569±27.32** | 151.18±139.13 | 178.91±150.19 | 133.78±101.91 |
| spleen | **26.62±43.77** | 53.64±88.81 | 79.88±108.99 | 48.17±85.19 |
| Pancreas | 82.63±101.68 | **73.82±81.91** | 91.99±117.98 | 75.79±77.21 |
| Aorta | **17.42±45.52** | 18.02±49.94 | 21.28±55.11 | 19.34±43.38 |
| Inferior Vena Cava | **24.84±36.42** | 57.01±87.62 | 68.99±85.14 | 51.33±55.41 |
| Right Adrenal Gland | **151.98±134.99** | 210.54±127.9 | 199.50±113.8 | 191.78±109.21 |
| Left Adrenal Gland | **150.53±108.66** | 160.10±112.73 | 177.98±113.42 | 167.18±115.67 |
| Gallbladder | 96.65±137.92 | 96.65±137.92 | 99.87±123.32 | **93.26±100.96** |
| Esophagus | **32.00±51.12** | 36.24±53.92 | 34.56±57.62 | 35.18±50.77 |
| heart | **19.23±29.71** | 23.15±32.29 | 45.61±50.19 | 21.89±36.79 |
| Left kidney | **31.18±75.36** | 41.66±65.18 | 69.81±96.18 | 47.81±73.29 |
| trachea | **5.86±8.46** | 9.38±43.4 | 19.18±44.81 | 10.46±18.90 |
| hippocampus | **28.00±14.13** | 84.96±102.63 | 140.71±100.91 | 88.91±90.79 |
| Stomach | **9.38±43.4** | 15.66±51.18 | 19.19±54.19 | 16.78±32.11 |
| hepatic vessel | **76.11±50.22** | 169.67±99.94 | 250.19±101.91 | 146.27±88.28 |
| colon | **105.94±139.81** | 107.93±143.28 | 152.89±128.98 | 125.21±134.26 |

Table C.C The complete comparison of *ASD* for four different methods

|  | TG-LMM (ours) | MedSAM | SAM | nnUnet |
|---|---|---|---|---|
| ALL | **15.21±28.92** | 24.41±33.35 | 48.55±54.11 | 90.25±125.22 |
| FLARE | **9.62±20.11** | 20.51±40.21 | 34.21±76.16 | 72.58±131.88 |
| segThor | **1.33±5.93** | 4.31±9.72 | 9.78±12.87 | 19.96±31.45 |
| MSD | **22.53±35.52** | 34.87±38.36 | 42.67±45.22 | 145.11±116.65 |
| liver | **18.19±24.43** | 44.87±37.27 | 79.89±65.32 | 47.11±39.13 |
| Right kidney | **1.037±5.691** | 33.15±35.23 | 61.23±61.78 | 34.81±37.00 |
| spleen | **5.26±6.41** | 11.36±19.20 | 21.12±35.96 | 11.93±20.16 |
| Pancreas | 20.67±25.79 | **19.42±18.51** | 33.78±34.87 | 20.39±19.44 |
| Aorta | 3.45±9.47 | **3.13±7.12** | 5.87±12.01 | 3.29±7.48 |
| Inferior Vena Cava | **5.1±6.85** | 11.21±15.40 | 21.04±26.73 | 11.77±16.17 |
| Right Adrenal Gland | **38.58±38.09** | 50.59±30.66 | 89.34±56.72 | 53.12±32.19 |
| Left Adrenal Gland | **33.72±25.66** | 40.50±31.85 | 71.85±55.21 | 42.53±33.44 |
| Gallbladder | **20.66±27.43** | 20.66±27.43 | 38.92±48.76 | 21.69±28.80 |
| Esophagus | **8.25±14.77** | 10.48±13.96 | 17.63±26.54 | 10.99±14.66 |
| heart | **3.75±4.89** | 4.95±5.98 | 8.01±11.87 | 5.20±6.28 |
| Left kidney | **6.89±20.17** | 12.34±27.78 | 23.97±48.53 | 12.96±29.17 |
| trachea | **1.38±2.98** | 1.97±10.06 | 3.21±19.02 | 2.07±10.56 |
| hippocampus | 5.10±6.12 | **4.65±3.12** | 7.89±6.01 | 4.88±3.28 |
| Stomach | **1.97±10.06** | 8.91±18.88 | 15.78±34.62 | 9.36±19.82 |
| hepatic vessel | **21.33±11.02** | 55.40±35.08 | 102.12±61.47 | 58.17±36.83 |
| colon | **28.52±36.84** | 31.41±45.49 | 57.98±79.84 | 33.00±47.76 |